\documentclass[preprint,12pt]{elsarticle}

\usepackage{amssymb}
\usepackage{graphicx,verbatim}
\usepackage{amsbsy}

\usepackage{amsmath}
\usepackage{booktabs}
\usepackage{multirow}
\usepackage{amssymb}
\usepackage{array}
\usepackage{makecell}
\usepackage{algorithm}
\usepackage{algpseudocode} 
\usepackage{placeins}
\usepackage{url}


\usepackage{lineno}
\graphicspath{{../}}

\journal{Computer Methods and Programs in Biomedicine}

\begin{document}

\begin{frontmatter}

\title{Angular Gaussian Supervised Contrastive Learning for Long-Tailed Electrocardiogram Arrhythmia Diagnosis}

\author[SJTU]{Jin Dai\fnref{associated}}

\author[TJ]{Qiuzhen Zhang\fnref{associated}}

\author[SJTU]{Chenyun Dai }

\author[TJ]{Danmei Lan \corref{cor1}}
\ead{landanmei2013@163.com}

\author[SJTU]{Can Han \corref{cor1}}
\ead{hancan@sjtu.edu.cn}

\fntext[associated]{Jin Dai and Qiuzhen Zhang contributed equally to this work as co-first authors.}

\cortext[cor1]{Corresponding author.}

\nonumnote{\raggedright
Our code is available at:
\url{https://github.com/Open-EXG/AG-SCL-for-Long-Tailed-ECG}.
}

\affiliation[SJTU]{organization={School of Biomedical Engineering, Shanghai Jiao Tong University},
            city={Shanghai},
            postcode={200030}, 
            country={China}}
\affiliation[TJ]{organization={Department of Neurology and Neurological Rehabilitation, Shanghai Yangzhi Rehabilitation Hospital, School of Medicine, Tongji University},
            city={Shanghai},
            postcode={201619}, 
            country={China}}

\begin{abstract}

\textbf{Background and Objective:}
Long-tailed label distributions limit the reliability of deep learning models for electrocardiogram (ECG) arrhythmia diagnosis, particularly for clinically important but low-prevalence rhythm abnormalities. 
Existing re-balancing and logit-adjustment methods usually treat rarity as the main source of difficulty, although ECG classes can also differ through direction-dependent morphological variability. 
This study proposes Angular Gaussian Supervised Contrastive Learning (AG-SCL) to improve long-tailed multi-label ECG diagnosis.

\textbf{Methods:}
AG-SCL combines three components within a shared ECG classification framework. 
First, an Angular Gaussian contrastive branch models full-covariance class uncertainty on unit-normalized embeddings, allowing direction-dependent feature variation to influence class separation.
Second, Adaptive Logit Adjustment learns bounded label-state-specific prior corrections instead of applying a fixed frequency-derived margin. 
Third, tail-aware augmentation generates morphology-preserving views while protecting the 7--25\,Hz QRS-dominant frequency band. 
The method was evaluated on the public PTB-XL benchmark and a collected nocturnal ECG (Noc-ECG) dataset comprising 1{,}317 hours from 141 subjects.

\textbf{Results:}
AG-SCL achieved the best macro performance on both datasets. 
On PTB-XL, it reached a balanced accuracy of 0.838, sensitivity of 0.709, specificity of 0.968, mean average precision of 0.495, and $TPR@FPR_{5\%}$ of 0.778. 
On Noc-ECG, it achieved a balanced accuracy of 0.918, sensitivity of 0.889, specificity of 0.947, mean average precision of 0.488, and $TPR@FPR_{5\%}$ of 0.900. 
The largest gains appeared in rare or unstable rhythm categories, and ablation analyses supported the contributions of the full Angular Gaussian covariance, Adaptive Logit Adjustment, and tail-aware augmentation.

\textbf{Conclusions:}
AG-SCL improves long-tailed ECG arrhythmia diagnosis by combining prior calibration with anisotropic representation learning. 
These results suggest that direction-aware contrastive modelling can improve rare-arrhythmia sensitivity while maintaining clinically relevant specificity constraints.

\end{abstract}


\end{frontmatter}


\section{Introduction}

Electrocardiogram (ECG) analysis is central to the non-invasive detection of clinically significant arrhythmias, from atrial fibrillation to ectopic rhythms \cite{ref_aha_ecg2009,ref_afib2024,ref_ecg_monitoring2017}. Wearable devices and long-term Holter monitoring now generate large volumes of rhythm data, increasing the need for reliable automated interpretation \cite{ref_rajpurkar2017,ref_perez2019}. Deep learning models have achieved strong performance on standard ECG benchmarks \cite{ref_atcnn,ref_t2t_vit,ref_vit_ecg,ref_liu2025}, but their reliability is reduced when diagnostic labels follow long-tailed distributions. In real ECG repositories, clinically important arrhythmias may account for less than $1\%$ of recordings \cite{ref_ptbxl_dataset,ref_deb2024,ref_ribeiro2020}, allowing common rhythms to dominate training and reduce sensitivity for rare diagnoses.

Prior work has addressed long-tailed ECG classification through stronger architectures, class re-balancing, loss calibration, representation learning, and data augmentation. These directions improve ECG classification from complementary angles, but they often handle data scarcity, feature geometry, and prior calibration separately. This leaves an unresolved gap for long-tailed multi-label ECG diagnosis, where rare-label performance depends on both limited positive observations and reliable decision calibration.

Representation-level learning provides a route to this gap by organizing class structure before the final classifier. Existing hyperspherical contrastive formulations often summarize class dispersion with isotropic uncertainty \cite{ref_supcon,ref_du2024,ref_hyperspherical_vae}, which has limited capacity to express direction-dependent ECG morphology when only a few positive examples are available.

Data augmentation can enrich scarce labels, but ECG perturbations must preserve diagnostically meaningful waveform structure. In particular, they should avoid corrupting the 7--25\,Hz band that carries much of the QRS information \cite{ref_thakor2007}.

\begin{figure}[t]
    \centering
    \includegraphics[width=\textwidth]{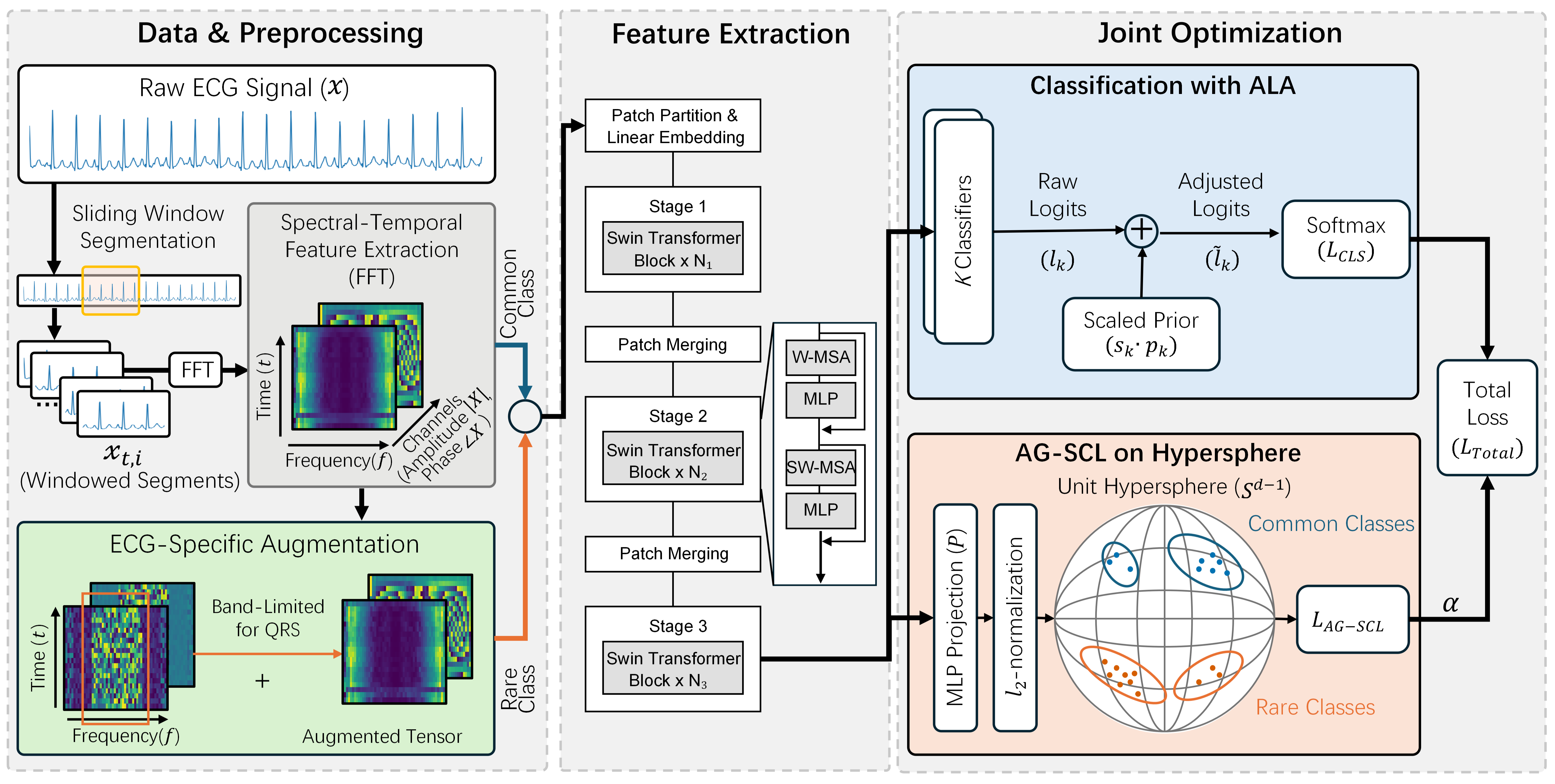}
    \caption{\textbf{Overview of AG-SCL.} ECG segments are converted to window-wise FFT magnitude--phase tensors, with band-limited augmentation (7--25\,Hz protected) applied to tail-label records during training. A Swin-Transformer backbone extracts features, followed by dual branches: one learns anisotropic Angular Gaussian representations on the unit hypersphere, and the other calibrates class priors via Adaptive Logit Adjustment (ALA).}
    \label{fig:main_fig}
    \end{figure}

To address these gaps, we propose \textbf{Angular Gaussian Supervised Contrastive Learning (AG-SCL)} for long-tailed multi-label ECG diagnosis (Fig.~\ref{fig:main_fig}). AG-SCL combines three components: tail-aware, band-constrained augmentation; an MGF-derived Angular Gaussian scoring model for direction-aware hyperspherical representations; and \textbf{Adaptive Logit Adjustment} (ALA) with label-state-specific prior calibration. Together, these components form a single training framework for sparse, morphology-variable ECG labels.
Our main contributions are:
\begin{itemize}
    \item We focus on a common but under-addressed setting in ECG diagnosis: long-duration rhythm monitoring in which clinically relevant arrhythmias are sparsely observed.

    \item We propose \textbf{Angular Gaussian Supervised Contrastive Learning (AG-SCL)}, which combines tail-aware augmentation with diagnostic-band constraints, Angular Gaussian representation learning, and \emph{Adaptive Logit Adjustment} for long-tailed multi-label ECG diagnosis.

    \item We establish Noc-ECG, a self-collected prospective nocturnal ECG cohort comprising 1{,}317 hours of overnight continuous lead-I recordings from 141 subjects, with expert-verified PAC/PVC labels and low positive ratios. We further evaluate AG-SCL on PTB-XL as an additional public validation benchmark.
\end{itemize}

\section{Related Work}

\subsection{Deep Learning for ECG Arrhythmia Diagnosis}

Deep learning has substantially advanced automated ECG interpretation by learning morphology and rhythm patterns directly from large-scale recordings. Early high-performing systems demonstrated cardiologist-level arrhythmia classification from ambulatory ECG and accurate diagnosis from 12-lead ECG recordings \cite{ref_rajpurkar2017,ref_ribeiro2020}. Public benchmarks such as PTB-XL further enabled standardized comparison across architectures, tasks, and diagnostic superclasses \cite{ref_ptbxl_paper,ref_strodthoff2021}. Subsequent ECG models have explored attention-based convolutional networks, transformer-style temporal modeling, and graph-based representations to improve feature extraction from variable-length or multi-lead signals \cite{ref_atcnn,ref_t2t_vit,ref_vit_ecg,ref_convrgnn}. These studies show that stronger architectures can improve overall ECG classification, but architecture design alone does not directly solve long-tailed label imbalance.

\subsection{Long-Tailed and Multi-Label Learning}

Long-tailed and multi-label learning methods address imbalance through data-level sampling, loss re-weighting, margin calibration, and classifier adjustment. Classical over-sampling methods such as SMOTE increase minority observations \cite{ref_smote}, whereas class-balanced loss estimates effective sample numbers to reduce head-class dominance \cite{ref_class_balanced_loss}. For multi-label recognition, distribution-balanced loss and asymmetric loss further consider label co-occurrence, positive-negative imbalance, and the large number of easy negative labels \cite{ref_distribution_balanced_loss,ref_asymmetric_loss}. Logit adjustment and recent long-tailed learning surveys show that prior-aware decision calibration is a principled route for improving imbalanced classifiers \cite{ref_logit_adj,ref_lt_survey}. However, these methods often connect correction strength to empirical frequency, which can be too rigid for heterogeneous ECG labels.

\subsection{Contrastive and Hyperspherical Representation Learning}

Supervised contrastive learning uses labels to pull together same-class samples and separate different-class samples \cite{ref_supcon}, and alignment-uniformity analysis explains why hyperspherical representations can support discriminative learning \cite{ref_align_uniform}. Long-tailed variants such as parametric contrastive learning and balanced contrastive learning explicitly modify contrastive objectives to reduce majority-class bias \cite{ref_paco,ref_bcl}. Probabilistic contrastive learning further models normalized class-conditional features with a von Mises--Fisher distribution \cite{ref_du2024}. These approaches improve long-tailed representation learning, but many probabilistic hyperspherical formulations summarize class dispersion with isotropic concentration parameters. This modeling choice limits direction-dependent uncertainty modeling.

\subsection{ECG and Time-Series Data Augmentation}

Generic approaches such as MixUp and CutMix regularize classifiers by interpolating samples or replacing local regions \cite{ref_mixup,ref_cutmix}. Broader time-series augmentation surveys also show that magnitude, time-domain, and frequency-domain perturbations can improve neural time-series classifiers \cite{ref_time_series_aug_survey,ref_iwana2021}. For ECG, however, augmentation must preserve diagnostic waveform structure rather than only increase distributional variety. Signal-processing studies indicate that the QRS complex carries substantial energy in the 7--25\,Hz range \cite{ref_thakor2007}, so unconstrained frequency perturbations may alter clinically important morphology. Existing augmentation strategies rarely combine long-tail targeting with explicit diagnostic-band protection.

\section{Method}

\subsection{Problem Formulation and Framework Overview}

We consider long-tailed multi-label ECG classification as a collection of $K$ related binary tasks. Given a training set $\mathcal{D}=\{(x_i,\mathbf{y}_i)\}_{i=1}^{N}$, each preprocessed single-channel 10s segment is $x_i\in\mathbb{R}^{L}$ ($L{=}1000$), and $\mathbf{y}_i=[y_{i,1},\ldots,y_{i,K}]^\top\in\{0,1\}^{K}$. For label $k$, the binary state is $c\in\{0,1\}$ and its empirical training prior is $\pi_{k,c}=N^{-1}\sum_i\mathbf{1}[y_{i,k}=c]$. We use \emph{tail label} to denote a diagnostic label whose positive state has low prevalence in the learning dataset.

As illustrated in Fig.~\ref{fig:main_fig}, AG-SCL uses one unmodified view for classification and two independently augmented views for contrastive representation learning. The three views share the encoder $F$ and are connected to label-specific classification and projection heads. ALA provides the state-specific prior correction used by both the classification and contrastive objectives. At inference, only the unmodified input, shared encoder, and binary classification heads are retained.

\subsection{Spectro-Temporal Input Representation}

AG-SCL operates on a window-wise spectro-temporal representation rather than on a single global Fourier transform. 
Each 10s segment $x_i$ contains $L{=}1000$ samples after preprocessing. We split it into 50 contiguous windows and compute a discrete Fourier transform within each window. 
The FFT magnitude and phase coefficients are retained as two channels. The resulting unmodified magnitude--phase tensor is denoted by $X_i^{(0)}$.

\subsection{Tail-Aware Multi-View Augmentation}\label{sec:augmentation}

During training, AG-SCL samples two independent stochastic views, $X_i^{(1)}$ and $X_i^{(2)}$, from each record. Let $\mathcal{T}\subseteq\{1,\ldots,K\}$ denote the tail-label set. A record is eligible for augmentation when $\sum_{k\in\mathcal{T}} y_{i,k}>0$; otherwise both stochastic views are left unchanged. For each eligible view, one candidate transformation is sampled with its own application parameter distribution. The original label vector is copied to every view.

The candidate pool contains six transformations. The first four are signal negation, $x'(t)=-x(t)$; global amplitude scaling, $x'(t)=g\,x(t)$ with $g\sim\mathcal{N}(1.0,0.2^2)$; additive jitter, $x'(t)=x(t)+\epsilon(t)$ with $\epsilon(t)\sim\mathcal{N}(0,\sigma_j^2)$ and $\sigma_j$ set to 1\% of the signal range; and frequency masking of a contiguous spectral interval capped at 30\% of the available spectral width.

The remaining two transformations are band-limited. Phase jitter uses $\phi'(f)=\phi(f)+\Delta\phi(f)$ with $\Delta\phi(f)\sim\mathcal{U}[-\delta_\phi(f),\delta_\phi(f)]$, and magnitude scaling uses $A'(f)=r(f)A(f)$. Their frequency-dependent ranges are
\[
\delta_\phi(f)=5^\circ\ (0.5\leq f<7),\quad 2^\circ\ (7\leq f\leq25),\quad 10^\circ\ (25<f\leq50),
\]
\[
\begin{aligned}
r(f)\sim{}&\mathcal{U}(0.90,1.10)\ (0.5\leq f<7),\quad \mathcal{U}(0.97,1.03)\ (7\leq f\leq25),\\
&\mathcal{U}(0.60,1.60)\ (25<f\leq50),
\end{aligned}
\]
where $f$ is measured in Hz. These settings apply the weakest perturbations in the 7--25\,Hz QRS-dominant band to preserve ventricular morphology while increasing tail-label view diversity.

\subsection{Shared Feature Extractor and Label-Specific Heads}\label{sec:backbone}

The three views share the same Swin-Transformer encoder $F$~\cite{ref_swin} during training. Patch embedding operates on the interval--frequency plane, and shifted-window attention connects neighboring temporal intervals and spectral regions. This makes the architecture suitable for the local and repeated nature of rhythm morphology. We use the hierarchical Swin stages shown in Fig.~\ref{fig:main_fig} as encoder model.

For view $v\in\{0,1,2\}$, the shared representation is $h_i^{(v)}=F(X_i^{(v)})\in\mathbb{R}^{D}$. The model then uses a separate binary classifier and projection head for every label. The original view produces raw classification logits $\boldsymbol{\ell}_{i,k}=G_k(h_i^{(0)})\in\mathbb{R}^{2}$. The augmented views are mapped by $P_k$ and normalized as
\begin{equation}
z_{i,k}^{(v)}=\frac{P_k(h_i^{(v)})}{\|P_k(h_i^{(v)})\|_2}
\in\mathcal{S}^{d-1},\qquad v\in\{1,2\}.
\label{eq:view-heads}
\end{equation}
Sharing $F$ transfers information across rhythm tasks, whereas $G_k$ and $P_k$ allow each label to retain its own binary decision space and positive/negative feature statistics. The unit-norm constraint makes similarity angular and prevents feature magnitude from dominating the class model.

This separation also assigns a clear role to each view. The raw classifier is optimized on $h_i^{(0)}$, matching the unmodified input available at deployment. In contrast, $P_k$ receives only the two stochastic training views and supplies the features from which class moments and contrast scores are computed. Gradients from both objectives meet in the shared encoder, allowing the distribution-aware branch to regularize the deployable representation without requiring its projection heads at test time.

\subsection{MGF-Derived Angular Gaussian Scoring}\label{sec:feature-distribution}

\subsubsection{From Isotropic Prototypes to Direction-Dependent Uncertainty}

ProCo models class-conditional features with an isotropic vMF distribution~\cite{ref_du2024}. To allow state-specific uncertainty to vary by direction, we associate every state $c$ of label $k$ with a random class prototype $W_{k,c}\sim\mathcal{N}(\boldsymbol{\mu}_{k,c},\boldsymbol{\Sigma}_{k,c})$ and marginalize it inside the exponential contrast kernel.

For a normalized query $z$, the Gaussian moment-generating function (MGF) gives
\begin{equation}
\begin{aligned}
q_{k,c}(z)
&=\log\mathbb{E}_{W_{k,c}}
  \left[\exp\!\left(\frac{z^\top W_{k,c}}{\tau}\right)\right] \\
&=\underbrace{\frac{z^\top\boldsymbol{\mu}_{k,c}}{\tau}}_{\text{prototype alignment}}
+\underbrace{\frac{z^\top\boldsymbol{\Sigma}_{k,c}z}{2\tau^2}}_{\text{direction-dependent uncertainty}},
\end{aligned}
\label{eq:mgf-score}
\end{equation}
where $\tau$ is the contrastive temperature. We refer to this component as \emph{Angular Gaussian} scoring because Gaussian prototype uncertainty is evaluated against angular, unit-normalized queries. If $\boldsymbol{\Sigma}_{k,c}=\sigma_{k,c}^{2}\mathbf{I}$, the quadratic term is direction invariant because $\|z\|_2=1$; a full covariance makes the score depend on the query direction and the class-specific principal axes.

\begin{figure}[t]
\centering
\includegraphics[width=0.7\textwidth]{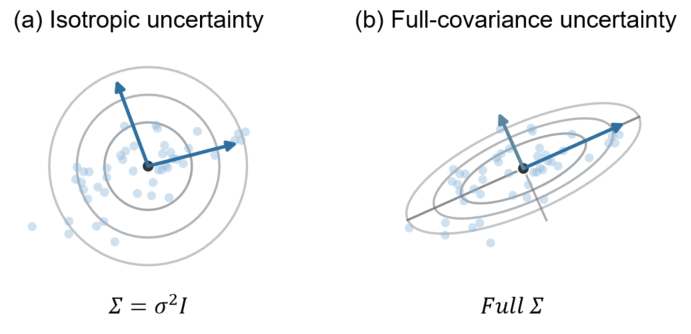}
\caption{
Conceptual comparison between isotropic and full-covariance class uncertainty.
\textbf{a}, An isotropic vMF model uses direction-invariant dispersion around the class mean.
\textbf{b}, The Angular Gaussian model uses a full covariance, allowing uncertainty to vary with the query direction and class-specific principal axes.
}
\label{fig:method_geometry}
\end{figure}

\subsubsection{Online Moment Estimation and Count-Based Shrinkage}

Maintaining class statistics online avoids storing a feature bank and allows the model to follow the evolving encoder. For the samples in the current mini-batch whose label-$k$ state is $c$, let $b_{k,c}^{(t)}$ be their count, $\bar z_{k,c}^{(t)}$ their mean, and $\bar Q_{k,c}^{(t)}$ the mean outer product $zz^\top$. With the accumulated count $n_{k,c}^{(t)}=n_{k,c}^{(t-1)}+b_{k,c}^{(t)}$ and data-dependent weight $\rho_{k,c}^{(t)}=b_{k,c}^{(t)}/n_{k,c}^{(t)}$, the first and second moments are updated by
\begin{equation}
\begin{aligned}
\boldsymbol{\mu}_{k,c}^{(t)}
&=(1-\rho_{k,c}^{(t)})\boldsymbol{\mu}_{k,c}^{(t-1)}
  +\rho_{k,c}^{(t)}\bar z_{k,c}^{(t)},\\
\mathbf{S}_{k,c}^{(t)}
&=(1-\rho_{k,c}^{(t)})\mathbf{S}_{k,c}^{(t-1)}
  +\rho_{k,c}^{(t)}\bar Q_{k,c}^{(t)},&
\boldsymbol{\Sigma}_{k,c}^{\mathrm{raw}}
&=\mathbf{S}_{k,c}^{(t)}-
  \boldsymbol{\mu}_{k,c}^{(t)}\boldsymbol{\mu}_{k,c}^{(t)\top}.
\end{aligned}
\label{eq:online-moments}
\end{equation}
These statistics are detached from gradient computation: they summarize the representation history, while gradients flow through the current query features and network parameters.

Both augmented views contribute to the online estimator. If a state is absent from a mini-batch, its update weight is zero and its stored moments remain unchanged. The accumulated count $n_{k,c}$ is then used to control shrinkage.

Direct full-covariance estimates are unreliable for a state observed only a few times. We therefore introduce an evidence-dependent interpolation with an isotropic identity prior:
\begin{equation}
\lambda_{k,c}^{(t)}=\exp\!\left(-\frac{n_{k,c}^{(t)}}{\tau_s}\right),\qquad
\widehat{\boldsymbol{\Sigma}}_{k,c}^{(t)}
=(1-\lambda_{k,c}^{(t)})\boldsymbol{\Sigma}_{k,c}^{\mathrm{raw}}
+\lambda_{k,c}^{(t)}\mathbf{I},
\label{eq:count-shrinkage}
\end{equation}
where $\tau_s$ controls how rapidly data replace the prior. At small counts, $\lambda_{k,c}^{(t)}$ is close to one and keeps the estimate near the identity prior. As evidence accumulates, it decays toward zero and allows the empirical covariance to dominate.

\subsection{Adaptive Logit Adjustment}\label{sec:classification}

ALA adds the log prior $\log\pi_{k,c}$ to state $c$, which assigns a bounded learnable scale to each label state and applies the resulting correction to both the classification and contrastive objectives:
\begin{equation}
\begin{gathered}
s_{k,c}=L+(H-L)\sigma(\gamma_{k,c}),\qquad
a_{k,c}=s_{k,c}\log\pi_{k,c},\\
\widetilde{\ell}_{i,k,c}=\ell_{i,k,c}+a_{k,c},\qquad
\widetilde{q}_{i,k,c}^{(v)}=q_{k,c}(z_{i,k}^{(v)})+a_{k,c}.
\end{gathered}
\label{eq:ala-both-branches}
\end{equation}
We set $L{=}0.5$ and $H{=}3.0$. Each $\gamma_{k,c}$ is initialized so that $s_{k,c}=1$, recovering fixed LA at the start of training. 
The bounded parameterization limits excessive prior correction, while the shared $a_{k,c}$ keeps the two objectives calibrated with the same state-specific prior term. 
At inference, evaluation uses the raw logits.

\subsection{Training Objective and Inference}\label{sec:optimization}

Let $y_{i,k}$ denote the observed binary state. The original-view classification loss and the two-view AG-SCL loss are
\begin{equation}
\begin{aligned}
\mathcal{L}_{\mathrm{CLS}}
&=-\frac{1}{N}\sum_{i=1}^{N}\sum_{k=1}^{K}
\log\frac{\exp(\widetilde{\ell}_{i,k,y_{i,k}})}
{\sum_{c=0}^{1}\exp(\widetilde{\ell}_{i,k,c})},\\
\mathcal{L}_{\mathrm{AG\text{-}SCL}}
&=-\frac{1}{2N}\sum_{v=1}^{2}\sum_{i=1}^{N}\sum_{k=1}^{K}
\log\frac{\exp(\widetilde{q}_{i,k,y_{i,k}}^{(v)})}
{\sum_{c=0}^{1}\exp(\widetilde{q}_{i,k,c}^{(v)})},\\
\mathcal{L}_{\mathrm{total}}
&=\mathcal{L}_{\mathrm{CLS}}+\alpha\mathcal{L}_{\mathrm{AG\text{-}SCL}}.
\end{aligned}
\label{eq:unified-objective}
\end{equation}
Here $\alpha$ balances discriminative classification and distribution-aware representation learning. Both losses use the shared ALA correction from Eq.~\eqref{eq:ala-both-branches}, but differ in their input views and logits.

Loss normalization follows the implementation: each binary loss is averaged over the mini-batch, the $K$ label losses are summed, and the two contrastive views are averaged. This keeps the relative weight $\alpha$ independent of the number of views.
Algorithm~\ref{alg:ag-scl} summarizes the complete training procedure.
During inference, predictions are obtained as $G_k(F(X^{(0)}))$ for every label. No augmented views are generated, class moments are not updated, and neither $P_k$ nor the MGF score is evaluated.

\begin{algorithm}[t]
\caption{Training AG-SCL for multi-label ECG diagnosis}
\label{alg:ag-scl}
\begin{algorithmic}[1]
\Require Mini-batch $\mathcal{B}$, priors $\{\pi_{k,c}\}$, encoder $F$, heads $\{G_k,P_k\}$, $\tau$, $\alpha$
\State Construct original spectral view $X^{(0)}$ and two independent eligible views $X^{(1)},X^{(2)}$
\State Encode all views: $h^{(v)}\gets F(X^{(v)})$, $v\in\{0,1,2\}$
\State $\mathcal{L}_{\mathrm{CLS}}\gets0$; $\mathcal{L}_{\mathrm{AG\text{-}SCL}}\gets0$
\For{$k=1$ to $K$}
    \State $\boldsymbol{\ell}_{k}\gets G_k(h^{(0)})$
    \State $z_k^{(v)}\gets\operatorname{normalize}(P_k(h^{(v)}))$ for $v\in\{1,2\}$
    \For{$v\in\{1,2\}$ and $c\in\{0,1\}$}
        \State Update $n_{k,c}$, $\boldsymbol{\mu}_{k,c}$, and $\mathbf{S}_{k,c}$ using state-$c$ features
        \State Form $\widehat{\boldsymbol{\Sigma}}_{k,c}$ by count-based shrinkage
        \State Compute $q_{k,c}^{(v)}$ with the MGF-derived score in Eq.~\eqref{eq:mgf-score}
    \EndFor
    \State Compute $s_{k,c}$ and adjust both $\ell_{k,c}$ and $q_{k,c}^{(v)}$
    \State Accumulate the original-view classification loss and two-view AG-SCL loss
\EndFor
\State $\mathcal{L}_{\mathrm{total}}\gets\mathcal{L}_{\mathrm{CLS}}+\alpha\mathcal{L}_{\mathrm{AG\text{-}SCL}}$
\State Update $F$, $\{G_k,P_k\}$, and $\{\gamma_{k,c}\}$ by back-propagation
\end{algorithmic}
\end{algorithm}

\section{Experimental Setup}
\subsection{Datasets and Cohort Construction}
We evaluated AG-SCL on PTB-XL and an in-house nocturnal ECG cohort (Noc-ECG), using ECG lead~I in both datasets. 
Here, a \emph{tail label} denotes a diagnostic label with low positive prevalence in the learning dataset. 
Table~\ref{tab:dataset_dist} summarizes the resulting label distributions.

\noindent\textbf{PTB-XL.}\quad
PTB-XL~\cite{ref_ptbxl_dataset,ref_ptbxl_paper} contains 10s, 12-lead clinical ECG recordings. 
We consolidated its rhythm annotations into six categories: AFIB/AFLT, PSVT/SVTAC, BIGU/TRIGU, STACH, SBRAD, and PACE. 
The official patient-wise split was used, with folds 1--8 for training, fold 9 for validation, and fold 10 for testing.

\noindent\textbf{Noc-ECG.}\quad
Noc-ECG comprised overnight lead-I recordings from adults with neurological disorders or high cardiovascular risk at Shanghai Yangzhi Rehabilitation Hospital, China. 
The demographic, anthropometric, and recording-environment characteristics of the 157 initially recruited participants are summarized in Table~\ref{tab:noc_cohort}. 
Each participant underwent one 8--10\,h overnight recording in a general ward using a clinical Holter recorder (BI9900, Biomedical Instruments, Shenzhen, China) at 500 or 1,000\,Hz. 
All recordings were resampled to 100\,Hz before segmentation and model input.

Quality control was performed before data partitioning. 
Sixteen subjects were excluded because of abnormal sampling frequencies ($n{=}3$), unrecovered electrode detachment ($n{=}8$), or more than 20\% unacceptable 10second windows according to the NeuroKit2 ECG-quality function ($n{=}5$)~\cite{Makowski2021neurokit}. 
The final cohort comprised 141 subjects and 1,317\,h of ECG, partitioned at the subject level into 83/29/29 subjects for training, validation, and testing. 
The target labels were premature atrial contractions (PAC) and premature ventricular contractions (PVC), both with positive ratios below 2\%. 
Initial beat-wise annotations generated by the Holter algorithm were manually verified by two cardiology experts, with disagreements resolved by consensus. 
A 10s segment was assigned a positive PAC or PVC label independently when it contained at least one verified abnormal beat.

\begin{table}[!htbp]
\centering
\caption{Positive sample ratios (\%) on PTB-XL and Noc-ECG.}
\label{tab:dataset_dist}
\small
\renewcommand{\arraystretch}{0.95}
\setlength{\tabcolsep}{4pt}
\begin{tabular}{@{}llr@{\hspace{0.8em}}llr@{}}
\toprule
\multicolumn{6}{@{}l}{\textit{PTB-XL}} \\
\cmidrule(lr){1-3}\cmidrule(l){4-6}
\textbf{ID} & \textbf{Rhythm label} & \textbf{Ratio (\%)} &
\textbf{ID} & \textbf{Rhythm label} & \textbf{Ratio (\%)} \\
\midrule
Cat 0 & AFIB/AFLT   & 7.47 & Cat 3 & PSVT/SVTAC & 0.20 \\
Cat 1 & STACH       & 3.94 & Cat 4 & BIGU/TRIGU & 0.27 \\
Cat 2 & SBRAD       & 3.04 & Cat 5 & PACE       & 1.40 \\
\midrule
\multicolumn{6}{@{}l}{\textit{Noc-ECG}} \\
\cmidrule(lr){1-3}\cmidrule(l){4-6}
\textbf{ID} & \textbf{Rhythm label} & \textbf{Ratio (\%)} &
\textbf{ID} & \textbf{Rhythm label} & \textbf{Ratio (\%)} \\
\midrule
PAC & \makecell[l]{Premature atrial\\contraction} & 1.98 &
PVC & \makecell[l]{Premature ventricular\\contraction} & 1.77 \\
\bottomrule
\end{tabular}
\end{table}

\begin{table}[!htbp]
\centering
\caption{Characteristics of the initially recruited Noc-ECG cohort ($N{=}157$). Continuous variables are mean $\pm$ standard deviation with range (minimum--maximum); statistics describe the pre-QC cohort.}
\label{tab:noc_cohort}
\small
\renewcommand{\arraystretch}{0.98}
\setlength{\tabcolsep}{3pt}
\begin{tabular}{@{}>{\raggedright\arraybackslash}p{0.28\linewidth}>{\raggedleft\arraybackslash}p{0.17\linewidth}@{\hspace{0.8em}}>{\raggedright\arraybackslash}p{0.25\linewidth}>{\raggedleft\arraybackslash}p{0.20\linewidth}@{}}
\toprule
\multicolumn{2}{@{}l}{\textit{Participants}} &
\multicolumn{2}{l}{\textit{Anthropometrics}} \\
\cmidrule(lr){1-2}\cmidrule(l){3-4}
\textbf{Characteristic} & \textbf{Value} &
\textbf{Characteristic} & \textbf{Value} \\
\midrule
Total, $N$ & 157 &
Age, years & $56.53 \pm 14.37$; 19--79 \\
Male, $N$ (\%) & 104 (66.24\%) &
Height, cm & $167.67 \pm 7.24$; 145--182 \\
Female, $N$ (\%) & 53 (33.76\%) &
Body mass index, kg/m$^2$ & \makecell[r]{$23.41 \pm 3.65$;\\12.17--36.38} \\
\midrule
\multicolumn{4}{@{}l}{\textit{Recording environment}} \\
\cmidrule(lr){1-2}\cmidrule(l){3-4}
Room temperature, $^{\circ}$C & $26.12 \pm 1.21$; 23.6--30.7 &
Relative humidity, \% & $65.89 \pm 6.75$; 46--81 \\
\bottomrule
\end{tabular}
\end{table}

\subsection{Preprocessing Protocol}
All recordings are resampled to 100\,Hz and divided into non-overlapping 10s segments. Each segment is bandpass-filtered between 3 and 40\,Hz and independently Z-score normalized. 

\subsection{Baselines and Comparison Protocol}
We compare AG-SCL with three ECG-specific architectures (AtCNN~\cite{ref_atcnn}, MTDL-NET~\cite{ref_t2t_vit}, and ViT-ECG~\cite{ref_vit_ecg}) and three long-tailed learning methods (GLA, GCA~\cite{ref_cortes2025}, and ProCo~\cite{ref_du2024}). 
All methods use identical patient-wise training, validation, and test partitions, ECG lead~I, and the same underlying samples and basic filtering pipeline.

The comparison follows a two-tier protocol. ECG-specific architectures retain training design prescribed by their original implementations, whereas the long-tailed methods are evaluated with a common input structure. 
This comparison therefore evaluates each complete method under its intended input design. 
Hyperparameters and checkpoints are selected exclusively on the validation set; the test set is used only for the final evaluation.

The ablation study uses two matched settings. The \emph{Factorial} setting varies the input domain, tail-aware augmentation, and learning objective. 
The \emph{Comp.} setting starts from the full AG-SCL model and removes the full Angular Gaussian covariance, ALA, or count-based shrinkage seperatedly.

\subsection{Evaluation Metrics}
Performance is evaluated at the segment level for each of the $K$ binary rhythm labels. Let $TP_k$, $TN_k$, $FP_k$, and $FN_k$ denote the confusion-matrix counts for label $k$. A fixed decision threshold of 0.5 is used for threshold-dependent metrics. We report the following measures:

\begin{itemize}
    \item \textbf{Sensitivity, Specificity, and Balanced Accuracy.}
    For label $k$,
    \begin{align}
        \mathrm{Sens}_k&=\frac{TP_k}{TP_k+FN_k}, &
        \mathrm{Spec}_k&=\frac{TN_k}{TN_k+FP_k}, &
        \mathrm{bACC}_k&=\frac{\mathrm{Sens}_k+\mathrm{Spec}_k}{2}.&&
    \end{align}

    \item \textbf{Average Precision and mean Average Precision.}
    \begin{equation}
        \mathrm{AP}_k=\sum_{n}\bigl(R_{k,n}-R_{k,n-1}\bigr)P_{k,n},
        \qquad
        \mathrm{mAP}=\frac{1}{K}\sum_{k=1}^{K}\mathrm{AP}_k,
    \end{equation}
    where $P_{k,n}$ and $R_{k,n}$ are the precision and recall at the $n^{th}$ operating point.

    \item \textbf{Operating-point metrics.}
    We also report two operating-point metrics:
    \begin{equation}
        P@R_{90,k}=\max_{t:\,R_k(t)\geq 0.90} P_k(t),
    \end{equation}
    \begin{equation}
        TPR@FPR_{5\%,k}=TPR_k(t_k^\star),
        \qquad
        t_k^\star=\arg\max_{t:\,FPR_k(t)\leq0.05} TPR_k(t).
    \end{equation}
    where $t$ ranges over the corresponding curve thresholds.
    
    \item \textbf{Range Precision, Range Recall, and Range-AUPRC.}
    Noc-ECG preserves continuous overnight timelines, enabling range-based evaluation unavailable for independent PTB-XL recordings. 
    Range recall measures recovery of true contiguous positive ranges; range precision measures how much predicted ranges match truth, penalizing fragmentation. 
    Consecutive positive 10s segments form ranges bounded by invalid segments, index gaps, or subject boundaries, and uniform positional weighting and reciprocal cardinality scores are averaged across ranges ~\cite{ref_tatbul2018}. 
    Let $\mathcal{R}=\{R_i\}_{i=1}^{N_R}$ denote ground-truth ranges and $\mathcal{P}=\{P_j\}_{j=1}^{N_P}$ predicted ranges:
    \begin{equation}
        \begin{gathered}
            m(A,\mathcal{B})
            =\left|\{B\in\mathcal{B}:A\cap B\neq\varnothing\}\right|,
            \enspace
            O(A,\mathcal{B})
            =\frac{\left|A\cap\left(\bigcup_{B\in\mathcal{B}}B\right)\right|}
            {|A|\max\{1,m(A,\mathcal{B})\}},\\
            \begin{aligned}
                \mathrm{RangeRecall}(\mathcal{R},\mathcal{P})
                &=\frac{1}{N_R}\sum_{i=1}^{N_R}O(R_i,\mathcal{P}),
                \quad N_R>0,\\
                \mathrm{RangePrecision}(\mathcal{R},\mathcal{P})
                &=\begin{cases}
                    \dfrac{1}{N_P}\displaystyle\sum_{j=1}^{N_P}O(P_j,\mathcal{R}), & N_P>0,\\
                    1, & N_P=0.
                \end{cases}
            \end{aligned}
        \end{gathered}
    \end{equation}
\end{itemize}

Except for mAP, the aggregate value of each metric $M$ is its unweighted macro average, $M_{macro}=K^{-1}\sum_{k=1}^{K}M_k$. We also report per-class Sensitivity, Specificity, AP, and $P@R_{90}$. For each run, the checkpoint with the highest validation mAP is used for test evaluation. Results are reported as the mean and standard deviation over five independent random seeds.

\subsection{Implementation Details}
The feature extractor is a frequency-domain Swin-style Transformer followed by label-specific heads that produce 256-dimensional, $\ell_2$-normalized representations for each binary label. Contrastive pairs are constructed independently for each label.
We train all models for 50 epochs with AdamW, a batch size of 256, an initial learning rate of $3{\times}10^{-4}$, and weight decay of $0.02$. The learning rate follows cosine annealing after a three-epoch warm-up. Unless otherwise stated, AG-SCL uses $\alpha{=}1$, $\tau{=}0.2$, and $\tau_s{=}800$.

\section{Results}

\subsection{Overall Results and Per-Class Diagnostic Performance}\label{sec:overall}

We benchmark AG-SCL against six representative baselines spanning two families: 
ECG-specific architectures (AtCNN~\cite{ref_atcnn}, MTDL-NET~\cite{ref_t2t_vit}, ViT-ECG~\cite{ref_vit_ecg}) and 
long-tailed learning methods (GLA and GCA~\cite{ref_cortes2025}, and ProCo~\cite{ref_du2024}). 

At the aggregate level, AG-SCL attains the best macro performance on both datasets (Table~\ref{tab:sota_comparison}). 
On PTB-XL it reaches a bACC of $0.838$, an mAP of $0.495$, and a $TPR@FPR_{5\%}$ of $0.778$; 
the gains remained strong on the clinically collected, long-duration Noc-ECG dataset, where both target labels had low positive ratios below 2\%. 
On this dataset, AG-SCL attains a bACC of $0.918$, an mAP of $0.488$, and a $TPR@FPR_{5\%}$ of $0.900$. 

\begin{figure}[!htbp]
\centering
\includegraphics[width=0.98\textwidth]{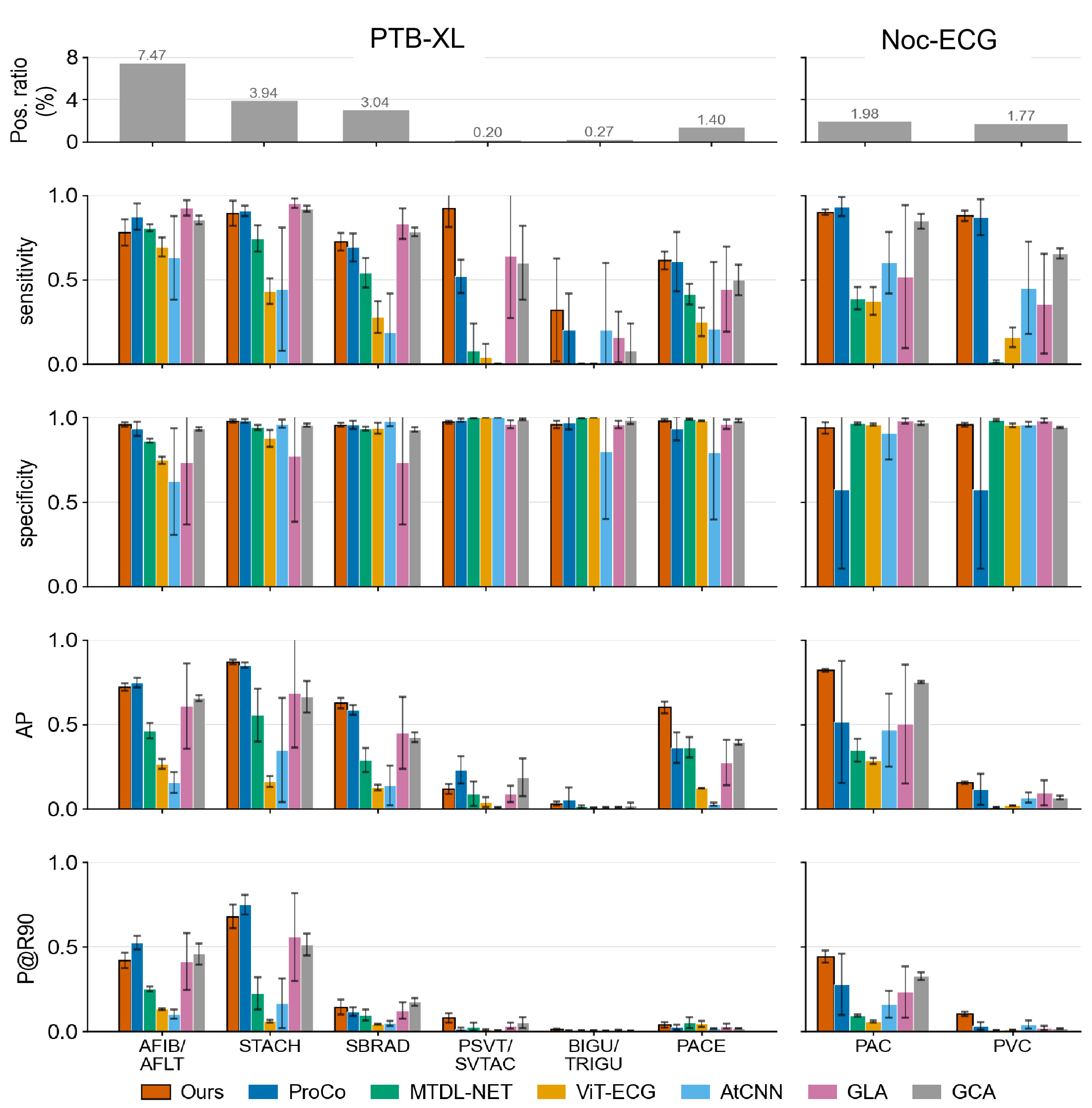}
\caption{Per-class performance on PTB-XL and Noc-ECG. The top row reports the per-class positive ratio (\%); the remaining rows report Sensitivity, Specificity, Average Precision, and $P@R_{90}$; error bars denote the standard deviation over five seeds.}
\label{fig:perclass}
\end{figure}

Fig.~\ref{fig:perclass} breaks performance down by category for all methods. 
On PTB-XL, the most pronounced advantage of AG-SCL appears on the rarest categories. 
For PSVT/SVTAC, AG-SCL recovers a sensitivity of $0.920$, far above the strongest baselines ProCo ($0.520$) and GCA ($0.600$); 
for the equally scarce BIGU/TRIGU ectopy (Cat~4, $0.27\%$) it still leads in sensitivity ($0.320$ vs.\ $0.200$ for ProCo and $0.080$ for GCA). 
On the more populated but still minority Pacemaker class (Cat~5), the improvement is also evident in ranking quality, with AG-SCL reaching an AP of $0.602$ versus $0.394$ (GCA) and $0.364$ (ProCo).

The same pattern holds on Noc-ECG. For PAC, AG-SCL attains an AP of $0.820$ (vs.\ $0.750$ for GCA) while maintaining a stable specificity of $0.938$; 
by contrast, ProCo shows a specificity of $0.575$ with very large variance. 
PVC remains the hardest class for all methods, yet AG-SCL still achieves the best AP ($0.156$). 
Overall, the per-class view shows that the macro gains are accompanied by stronger tail-class sensitivity and more stable ranking performance, rather than uniform improvements across all classes.

\subsection{Comparison with State-of-the-Art}\label{sec:sota}

Table~\ref{tab:sota_comparison} details the per-metric standings across all six baselines. On PTB-XL, AG-SCL outperforms the strongest baseline GCA on balanced accuracy ($+1.9\%$) and $TPR@FPR_{5\%}$ ($+5.8\%$), and additionally attains the best mAP ($0.495$ vs.\ ProCo's $0.471$). 

The advantage is more pronounced on the low-prevalence, long-duration Noc-ECG dataset. AG-SCL improves balanced accuracy by $+6.6\%$ over GCA, increases $TPR@FPR_{5\%}$ to $0.900$ (vs.\ $0.753$), and raises mAP to $0.488$ (vs.\ $0.409$). Its specificity is not the highest in the table, but it preserves a more balanced sensitivity-specificity profile than ProCo, which reaches the highest sensitivity ($0.902$) while dropping to a specificity of $0.575 \pm 0.469$. The smaller standard deviations of AG-SCL on Noc-ECG further show more stable behaviour across seeds than the contrastive and re-weighting baselines.

\begin{table}[!htbp]
\centering
\caption{SOTA comparison on PTB-XL and Noc-ECG datasets, shown as Mean $\pm$ Standard Deviation. The best-performing is in \textbf{bold}, and the second-best is \underline{underlined}.}
\label{tab:sota_comparison}
\resizebox{\textwidth}{!}{%
\begin{tabular}{llcccccc}
\toprule
\textbf{Dataset} & \textbf{Method} & \textbf{bACC} & \textbf{Sens} & \textbf{Spec} & \textbf{mAP} & \textbf{$P@R_{90}$} & \textbf{$TPR@FPR_{5\%}$} \\
\midrule
\multirow{7}{*}{PTB-XL} 
 & AtCNN \cite{ref_atcnn} & $0.569 \pm 0.060$ & $0.278 \pm 0.155$ & $0.859 \pm 0.180$ & $0.114 \pm 0.081$ & $0.057 \pm 0.031$ & $0.192 \pm 0.115$ \\
 & ViT-ECG \cite{ref_vit_ecg} & $0.603 \pm 0.011$ & $0.282 \pm 0.026$ & $0.923 \pm 0.007$ & $0.121 \pm 0.016$ & $0.049 \pm 0.004$ & $0.267 \pm 0.022$ \\
 & MTDL-NET \cite{ref_t2t_vit} & $0.693 \pm 0.014$ & $0.431 \pm 0.026$ & $0.954 \pm 0.004$ & $0.296 \pm 0.026$ & $0.109 \pm 0.024$ & $0.556 \pm 0.039$ \\
 & GLA \cite{ref_cortes2025} & $0.756 \pm 0.129$ & $0.658 \pm 0.085$ & $0.853 \pm 0.177$ & $0.353 \pm 0.158$ & $0.194 \pm 0.083$ & $0.549 \pm 0.248$ \\
 & ProCo \cite{ref_du2024} & $0.796 \pm 0.023$ & $0.633 \pm 0.075$ & $0.959 \pm 0.029$ & \underline{$0.471 \pm 0.037$} & \textbf{0.238 $\pm$ 0.013} & $0.684 \pm 0.020$ \\
 & GCA \cite{ref_cortes2025} & \underline{$0.819 \pm 0.013$} & \underline{$0.679 \pm 0.029$} & \underline{$0.959 \pm 0.006$} & $0.379 \pm 0.036$ & $0.217 \pm 0.025$ & \underline{$0.720 \pm 0.017$} \\
 & \textbf{Ours} & \textbf{0.838 $\pm$ 0.030} & \textbf{0.709 $\pm$ 0.069} & \textbf{0.968 $\pm$ 0.009} & \textbf{0.495 $\pm$ 0.009} & \underline{$0.229 \pm 0.018$} & \textbf{0.778 $\pm$ 0.034} \\
\midrule
\multirow{7}{*}{Noc-ECG} 
 & MTDL-NET \cite{ref_t2t_vit} & $0.588 \pm 0.017$ & $0.202 \pm 0.036$ & \underline{$0.975 \pm 0.006$} & $0.179 \pm 0.034$ & $0.051 \pm 0.003$ & $0.258 \pm 0.031$ \\
 & ViT-ECG \cite{ref_vit_ecg} & $0.611 \pm 0.023$ & $0.266 \pm 0.054$ & $0.956 \pm 0.009$ & $0.152 \pm 0.007$ & $0.033 \pm 0.003$ & $0.295 \pm 0.021$ \\
 & GLA \cite{ref_cortes2025} & $0.709 \pm 0.171$ & $0.438 \pm 0.358$ & \textbf{0.981 $\pm$ 0.015} & $0.299 \pm 0.213$ & $0.126 \pm 0.082$ & $0.500 \pm 0.368$ \\
 & AtCNN \cite{ref_atcnn} & $0.729 \pm 0.146$ & $0.526 \pm 0.226$ & $0.932 \pm 0.082$ & $0.267 \pm 0.123$ & $0.100 \pm 0.050$ & $0.624 \pm 0.310$ \\
 & ProCo \cite{ref_du2024} & $0.738 \pm 0.194$ & \textbf{0.902 $\pm$ 0.081} & $0.575 \pm 0.469$ & $0.315 \pm 0.227$ & $0.155 \pm 0.101$ & $0.527 \pm 0.398$ \\
 & GCA \cite{ref_cortes2025} & \underline{$0.852 \pm 0.012$} & $0.750 \pm 0.027$ & $0.954 \pm 0.004$ & \underline{$0.409 \pm 0.008$} & \underline{$0.171 \pm 0.011$} & \underline{$0.753 \pm 0.013$} \\
 & \textbf{Ours} & \textbf{0.918 $\pm$ 0.007} & \underline{$0.889 \pm 0.024$} & $0.947 \pm 0.022$ & \textbf{0.488 $\pm$ 0.008} & \textbf{0.273 $\pm$ 0.023} & \textbf{0.900 $\pm$ 0.010} \\
\bottomrule
\end{tabular}%
}
\end{table}

\subsection{Subject-Level and Range-Based Evaluation on Noc-ECG}\label{sec:subject_level}

\begin{figure}[!htbp]
\centering
\includegraphics[width=0.94\textwidth]{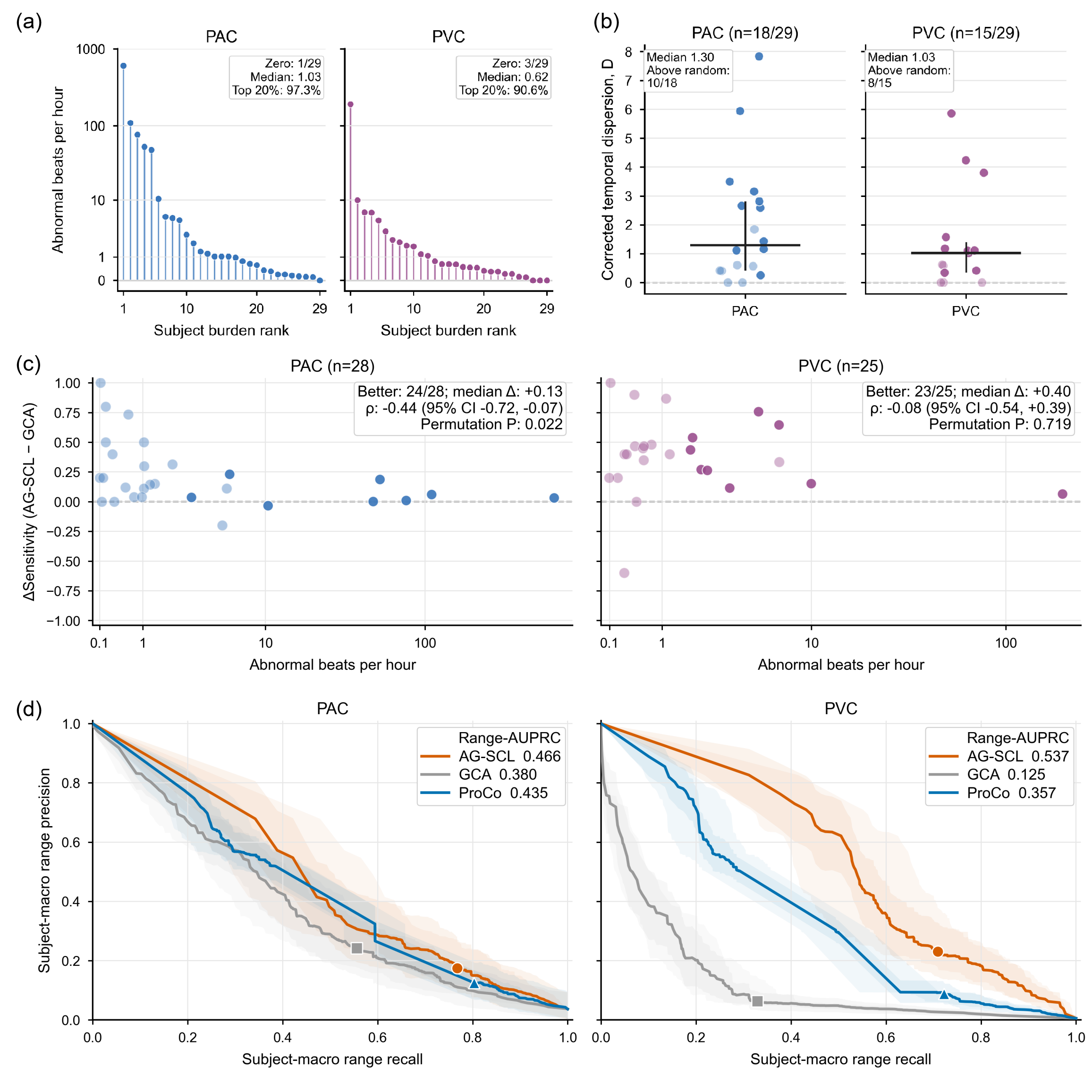}
\caption{Subject-level temporal structure and range-based performance on Noc-ECG. 
\textbf{(a)} Ranked PAC/PVC burden. 
\textbf{(b)} Corrected temporal dispersion across non-overlapping 30-min bins: $D=\log_2[F_{\mathrm{obs}}/\operatorname{median}(F_{\mathrm{random}})]$. 
\textbf{(c)} AG-SCL minus GCA sensitivity; lighter points indicate fewer than 10 positive segments. 
\textbf{(d)} Subject-macro range precision--recall curves with 95\% subject-bootstrap CIs; symbols mark threshold 0.5.}
\label{fig:subject_level}
\end{figure}

Across the 29 held-out Noc-ECG subjects, PAC and PVC burdens were highly uneven(Fig.~\ref{fig:subject_level}a).
Corrected temporal dispersion $D$ compared observed beat-count variability across 30-min bins with a within-subject randomized reference to test whether events clustered beyond chance. 
Fig.~\ref{fig:subject_level}b showed positive median $D$ for both PAC and PVC, with 10/18 and 8/15 eligible subjects. 
Together, these findings showed subject-level imbalance and non-random temporal clustering of PAC and PVC events in a subset of eligible Noc-ECG recordings.

We next compared AG-SCL with the most stable baseline(GCA) to determine whether the pooled sensitivity gain extended across subjects with at least one positive segment for each label (Fig.~\ref{fig:subject_level}c). 
AG-SCL achieved higher subject-level sensitivity in 24/28 PAC-positive subjects and 23/25 PVC-positive subjects. 
Sensitivity gains extended across the observed burden range, indicating that the model improvement was not confined to subjects with a particular burden level.

Range-based evaluation compared AG-SCL with GCA and ProCo(Fig.~\ref{fig:subject_level}d). 
Multi-segment ranges contained 61.5\% of PAC-positive and 40.0\% of PVC-positive segments. 
AG-SCL achieved the highest Range-AUPRC point estimate for both PAC (0.466 vs. 0.435 for ProCo and 0.380 for GCA) and PVC (0.537 vs. 0.357 and 0.125, respectively).

\subsection{Ablation Study}
\begin{figure}[!htbp]
\centering
\includegraphics[width=1\textwidth]{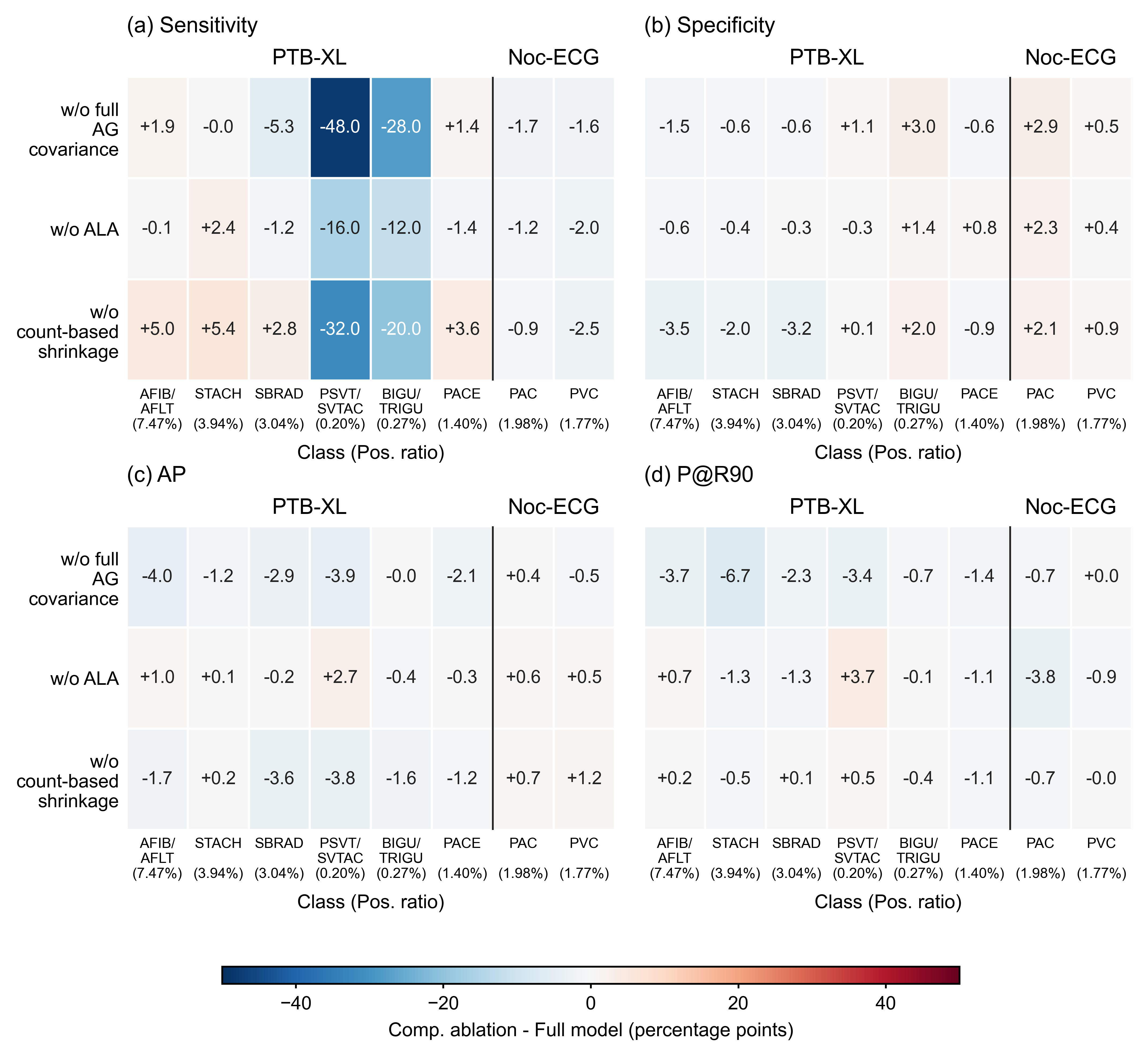}
\caption{Per-class component-ablation deltas relative to the full model. Values are ablation minus Full in percentage points, averaged over five seeds. Rows correspond to the three Component settings in Table~\ref{tab:ablation}; columns show PTB-XL and Noc-ECG classes with their positive ratios. Blue cells denote performance loss after removing a module, and red cells denote a gain.}
\label{fig:ablation_component_perclass}
\end{figure}

The ablation study showed that AG-SCL benefited from the FFT-phase input, the contrastive learning objective, and the three component modules (Table~\ref{tab:ablation}, Fig.~\ref{fig:ablation_component_perclass}).

\noindent\textbf{Factorial design.}\quad
As shown in Table~\ref{tab:ablation}, replacing AG-SCL with cross-entropy under the same FFT-phase and augmentation setting reduced PTB-XL bACC by $5.5$ points and mAP by $10.4$ points.
Replacing FFT Phase with raw time input reduced bACC from $83.8$ to $77.7$ and increased its standard deviation from $3.0$ to $9.1$ points.
Removing rare-class augmentation produced a smaller decline, from $83.8$ to $80.0$ bACC and from $77.8$ to $73.7$ $TPR@FPR_{5\%}$.
On Noc-ECG, the same factorial ordering was observed, with smaller absolute drops than on PTB-XL.

\noindent\textbf{Component ablation.}\quad
The component ablations caused larger aggregate losses on PTB-XL.
Removing the full AG covariance reduced bACC from $83.8$ to $77.4$ and sensitivity from $70.9$ to $57.9$.
Removing ALA caused a smaller PTB-XL decline, whereas removing count-based shrinkage increased seed-to-seed variability in sensitivity ($65.0_{\pm12.5}$).
On Noc-ECG, the component losses were smaller but still visible in sensitivity-related metrics.
Sensitivity decreased from $88.9$ to $87.2$--$87.3$ after removing any component, and $P@R_{90}$ decreased from $27.3$ to $25.0$--$27.0$.
$TPR@FPR_{5\%}$ decreased after removing ALA or count-based shrinkage, while specificity increased across the three component ablations.
In the per-class deltas, PTB-XL losses were concentrated in the rare PSVT/SVTAC and BIGU/TRIGU classes, whereas Noc-ECG showed narrower changes across PAC and PVC (Fig.~\ref{fig:ablation_component_perclass}).

\begin{table}[!htbp]
\centering
\caption{Ablation study on PTB-XL and Noc-ECG, reported as mean$_{\pm\text{std}}$ (\%). The \emph{Factorial} block sweeps loss, input domain, and augmentation (a $\checkmark$ in the Aug column marks rare-class augmentation, applied only in the FFT Phase domain). The \emph{Comp.} block removes one AG-SCL module from the full model. Bold and underlined values mark the best and second-best settings within the Factorial block only.}
\label{tab:ablation}
\begingroup
\renewcommand{\arraystretch}{0.96}
\resizebox{0.94\textwidth}{!}{%
\begin{tabular}{llllcccccc}
\toprule
\textbf{Block} & \textbf{Loss} & \textbf{Domain} & \textbf{Aug} & \textbf{bACC} & \textbf{Sens} & \textbf{Spec} & \textbf{mAP} & \textbf{$P@R_{90}$} & \textbf{$TPR@FPR_{5\%}$} \\
\midrule
\multicolumn{10}{c}{\textit{PTB-XL}} \\
\midrule
\multirow{6}{*}{\rotatebox{90}{Factorial}}
 & AG-SCL & FFT Phase & $\checkmark$ \footnotesize(Full) & $\mathbf{83.8}_{\pm3.0}$ & $\mathbf{70.9}_{\pm6.9}$ & $96.8_{\pm0.9}$ & $\mathbf{49.5}_{\pm0.9}$ & $\mathbf{22.9}_{\pm1.8}$ & $\mathbf{77.8}_{\pm3.4}$ \\
 & CE     & FFT Phase & $\checkmark$ & $78.3_{\pm3.7}$ & $64.4_{\pm8.9}$ & $92.1_{\pm1.9}$ & $39.1_{\pm2.0}$ & $18.6_{\pm1.2}$ & $70.6_{\pm2.0}$ \\
 & AG-SCL & FFT Phase &              & $\underline{80.0}_{\pm2.6}$ & $62.9_{\pm6.4}$ & $\underline{97.2}_{\pm1.3}$ & $\underline{47.9}_{\pm0.8}$ & $\underline{22.3}_{\pm0.8}$ & $\underline{73.7}_{\pm2.7}$ \\
 & CE     & FFT Phase &              & $73.5_{\pm1.0}$ & $49.9_{\pm1.9}$ & $97.1_{\pm0.6}$ & $45.4_{\pm0.9}$ & $20.7_{\pm0.5}$ & $73.0_{\pm2.1}$ \\
 & AG-SCL & Time      &              & $77.7_{\pm9.1}$ & $\underline{64.8}_{\pm13.8}$ & $90.6_{\pm4.9}$ & $34.7_{\pm15.0}$ & $19.4_{\pm9.2}$ & $57.8_{\pm20.9}$ \\
 & CE     & Time      &              & $75.6_{\pm2.4}$ & $53.5_{\pm6.1}$ & $\mathbf{97.7}_{\pm1.4}$ & $46.0_{\pm3.2}$ & $21.4_{\pm1.6}$ & $68.9_{\pm3.7}$ \\
\cmidrule(lr){1-10}
\multirow{3}{*}{\rotatebox{90}{Comp.}}
 & \multicolumn{3}{l}{w/o full AG covariance} & $77.4_{\pm3.6}$ & $57.9_{\pm8.9}$ & $96.9_{\pm1.7}$ & $47.1_{\pm0.4}$ & $19.8_{\pm1.5}$ & $68.4_{\pm4.6}$ \\
 & \multicolumn{3}{l}{w/o ALA (fixed LA, $s_k{\equiv}1$)} & $81.5_{\pm4.0}$ & $66.1_{\pm9.4}$ & $96.9_{\pm1.5}$ & $49.9_{\pm2.4}$ & $23.0_{\pm1.0}$ & $74.1_{\pm4.7}$ \\
 & \multicolumn{3}{l}{w/o count-based shrinkage}          & $80.2_{\pm5.3}$ & $65.0_{\pm12.5}$ & $95.5_{\pm2.0}$ & $47.5_{\pm1.0}$ & $22.7_{\pm1.7}$ & $70.8_{\pm8.1}$ \\
\midrule
\multicolumn{10}{c}{\textit{Noc-ECG}} \\
\midrule
\multirow{6}{*}{\rotatebox{90}{Factorial}}
 & AG-SCL & FFT Phase & $\checkmark$ \footnotesize(Full) & $\mathbf{91.8}_{\pm0.7}$ & $\mathbf{88.9}_{\pm2.4}$ & $94.7_{\pm2.2}$ & $48.8_{\pm0.8}$ & $\mathbf{27.3}_{\pm2.3}$ & $\mathbf{90.0}_{\pm1.0}$ \\
 & CE     & FFT Phase & $\checkmark$ & $86.5_{\pm0.9}$ & $74.7_{\pm1.9}$ & $\mathbf{98.3}_{\pm0.1}$ & $47.8_{\pm0.5}$ & $23.5_{\pm1.5}$ & $87.1_{\pm0.8}$ \\
 & AG-SCL & FFT Phase &              & $\underline{91.4}_{\pm0.8}$ & $\underline{87.0}_{\pm2.0}$ & $95.8_{\pm0.9}$ & $\mathbf{49.8}_{\pm2.3}$ & $\underline{26.3}_{\pm2.7}$ & $89.2_{\pm1.1}$ \\
 & CE     & FFT Phase &              & $89.6_{\pm0.3}$ & $81.5_{\pm0.6}$ & $97.7_{\pm0.1}$ & $\underline{49.6}_{\pm0.8}$ & $25.8_{\pm2.1}$ & $\underline{89.8}_{\pm0.6}$ \\
 & AG-SCL & Time      &              & $88.6_{\pm0.9}$ & $80.7_{\pm1.8}$ & $96.6_{\pm0.4}$ & $49.3_{\pm1.8}$ & $23.7_{\pm3.3}$ & $85.2_{\pm1.7}$ \\
 & CE     & Time      &              & $84.0_{\pm1.4}$ & $70.1_{\pm3.0}$ & $\underline{97.9}_{\pm0.1}$ & $46.8_{\pm0.9}$ & $20.2_{\pm2.3}$ & $82.0_{\pm0.5}$ \\
\cmidrule(lr){1-10}
\multirow{3}{*}{\rotatebox{90}{Comp.}}
 & \multicolumn{3}{l}{w/o full AG covariance} & $91.8_{\pm0.2}$ & $87.2_{\pm0.8}$ & $96.4_{\pm0.5}$ & $48.7_{\pm1.0}$ & $27.0_{\pm1.2}$ & $90.0_{\pm0.5}$ \\
 & \multicolumn{3}{l}{w/o ALA (fixed LA, $s_k{\equiv}1$)} & $91.7_{\pm0.4}$ & $87.3_{\pm1.5}$ & $96.1_{\pm0.9}$ & $49.3_{\pm1.1}$ & $25.0_{\pm2.2}$ & $89.3_{\pm0.4}$ \\
 & \multicolumn{3}{l}{w/o count-based shrinkage}          & $91.7_{\pm0.4}$ & $87.2_{\pm1.3}$ & $96.2_{\pm0.6}$ & $49.7_{\pm0.8}$ & $26.9_{\pm1.2}$ & $89.8_{\pm0.4}$ \\
\bottomrule
\end{tabular}%
}
\endgroup
\end{table}

\subsection{Sensitivity Analysis}
\begin{figure}[!htbp]
\centering
\includegraphics[width=1\textwidth]{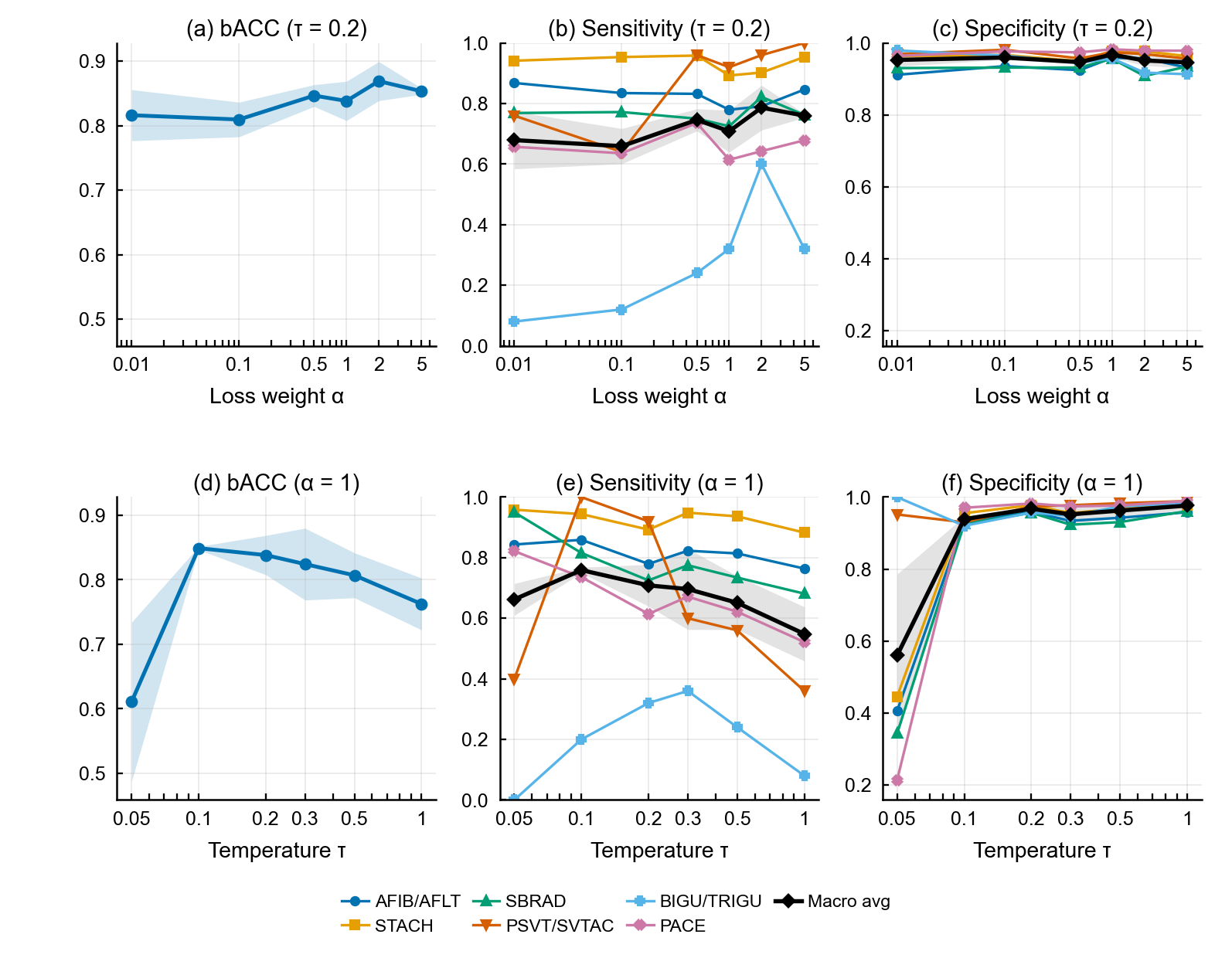}
\caption{Per-class hyperparameter sensitivity on PTB-XL (mean$\pm$std over five seeds). \textbf{Top row:} bACC, Sensitivity, and Specificity as a function of the loss weight $\alpha$ ($\tau{=}0.2$). \textbf{Bottom row:} the same metrics as a function of the temperature $\tau$ ($\alpha{=}1$). Coloured lines denote the six rhythm categories and the black line the macro average; shading denotes the standard deviation. Head classes stay flat, whereas the rarest categories (SVT, BIGU/TRIGU ectopy) swing widely and destabilize at small $\tau$.}
\label{fig:sensitivity_ptbxl}
\end{figure}

\begin{figure}[!htbp]
\centering
\includegraphics[width=1\textwidth]{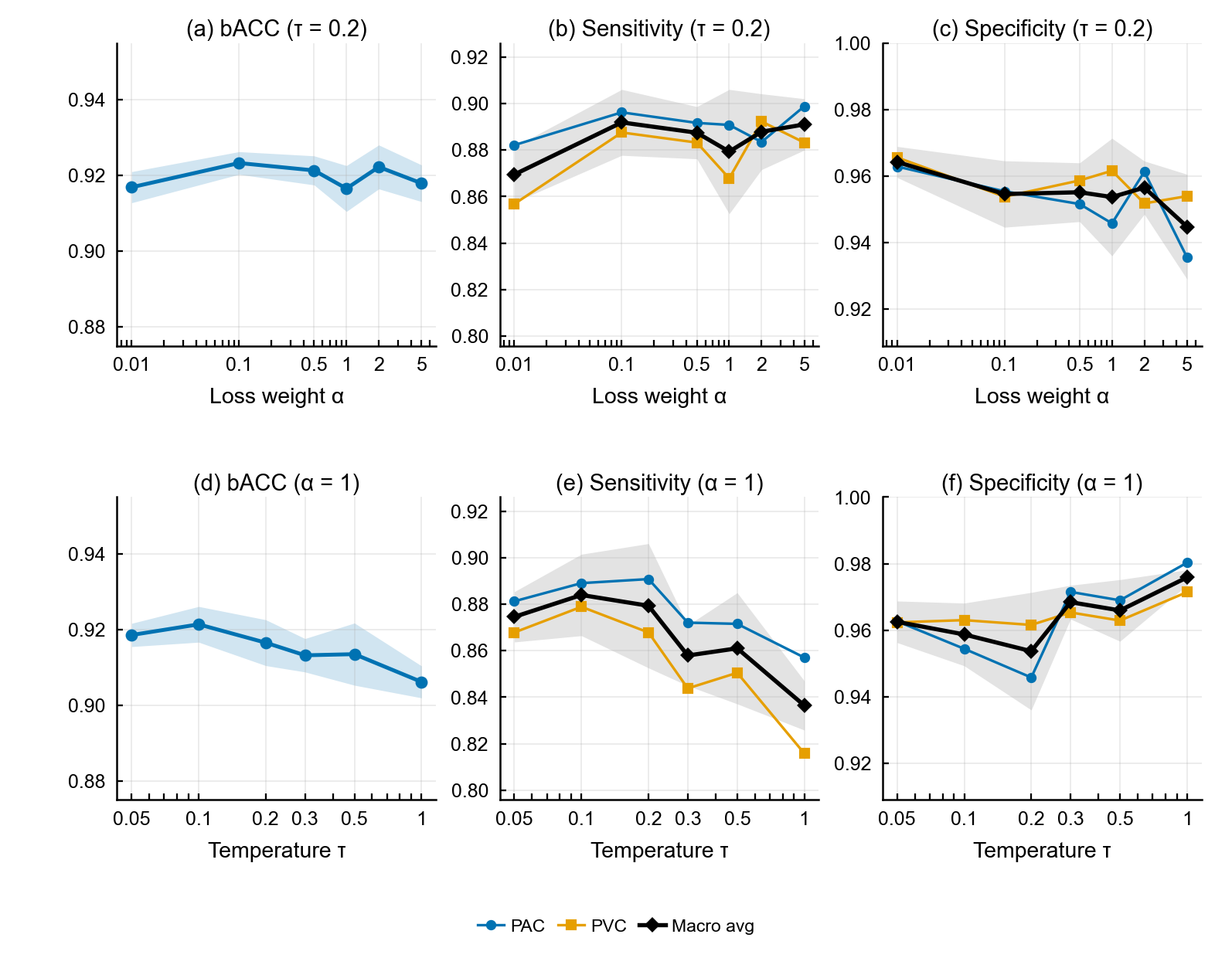}
\caption{Per-class hyperparameter sensitivity on Noc-ECG (mean$\pm$std over five seeds), with the same layout as Fig.~\ref{fig:sensitivity_ptbxl}: \textbf{top row} vs.\ the loss weight $\alpha$ ($\tau{=}0.2$) and \textbf{bottom row} vs.\ the temperature $\tau$ ($\alpha{=}1$). Coloured lines denote PAC and PVC and the black line the macro average. Both classes remain robust across the entire hyperparameter range.}
\label{fig:sensitivity_noc}
\end{figure}

We study the two principal hyperparameters of AG-SCL, the contrastive loss weight $\alpha$ (sweeping $\alpha$ at fixed $\tau{=}0.2$) and the temperature $\tau$ (sweeping $\tau$ at fixed $\alpha{=}1$), and resolve their effect not only on the macro average but on every rhythm category (Fig.~\ref{fig:sensitivity_ptbxl} for PTB-XL, Fig.~\ref{fig:sensitivity_noc} for Noc-ECG; all values are mean$\pm$std over five seeds).
At the macro level the model is insensitive to $\alpha$, peaking around $\alpha\!\in\![1.0,\,2.0]$, whereas $\tau$ exerts a markedly stronger influence on performance: an overly small $\tau{=}0.05$ destabilizes training on PTB-XL (macro bACC drops to $0.610$), and the optimum sits at $\tau\!\approx\!0.1$--$0.2$ on both datasets.
This sweet spot is shared across categories, so the macro-optimal setting does not trade head-class accuracy for tail-class recall.

The per-class breakdown shows that this aggregate robustness masks a strong dependence on class rarity.
On PTB-XL, the well-populated head classes are flat across both sweeps. SinusTachy (positive ratio $3.94\%$) and AF/Flutter ($7.47\%$) vary by only $\approx\!0.08$ and $\approx\!0.10$ in sensitivity over the entire $\tau$ range. In contrast, the rarest categories are highly volatile: PSVT/SVTAC (SVT, $0.20\%$) swings from a sensitivity of $0.36$ to $1.00$ across $\tau$, BIGU/TRIGU ectopy ($0.27\%$) from $0.00$ to $0.36$, and Pacemaker ($1.40\%$) from $0.52$ to $0.82$; the loss weight $\alpha$ induces a comparable spread on the tail (e.g.\ ectopy sensitivity $0.08\!\to\!0.60$).
The macro instability at $\tau{=}0.05$ is driven mainly by these tail classes. At the extreme temperature, the rare categories collapse and become high-variance across seeds, whereas the head-class curves remain smooth. By contrast, Noc-ECG is robust for both PAC ($1.98\%$) and PVC ($1.77\%$): sensitivity varies by less than $0.07$ and specificity by less than $0.04$ across every $\alpha$ and $\tau$ setting. Thus, the observed hyperparameter sensitivity arises primarily from the $<1\%$ PTB-XL tail categories.

\subsection{Empirical Validation of the Anisotropy Assumption}
To evaluate whether the anisotropic assumption reflected a reproducible property of the learned feature distributions, we performed a held-out tangent-space fitting analysis on frozen projection features. 
For each eligible label state, the samples were repeatedly divided into equal train and held-out subsets. The train subset was used to define the local tangent basis, estimate the mean and covariance, and construct two density surrogates: an isotropic vMF-like model and an anisotropic AG model. 
The held-out subset was used only for negative log-likelihood evaluation. We report $\Delta\mathrm{NLL}=\mathrm{NLL}_{\mathrm{isotropic}}-\mathrm{NLL}_{\mathrm{AG}}$, where positive values indicate a lower held-out NLL for the anisotropic covariance.

\begin{figure}[!htbp]
\centering
\includegraphics[width=1\textwidth]{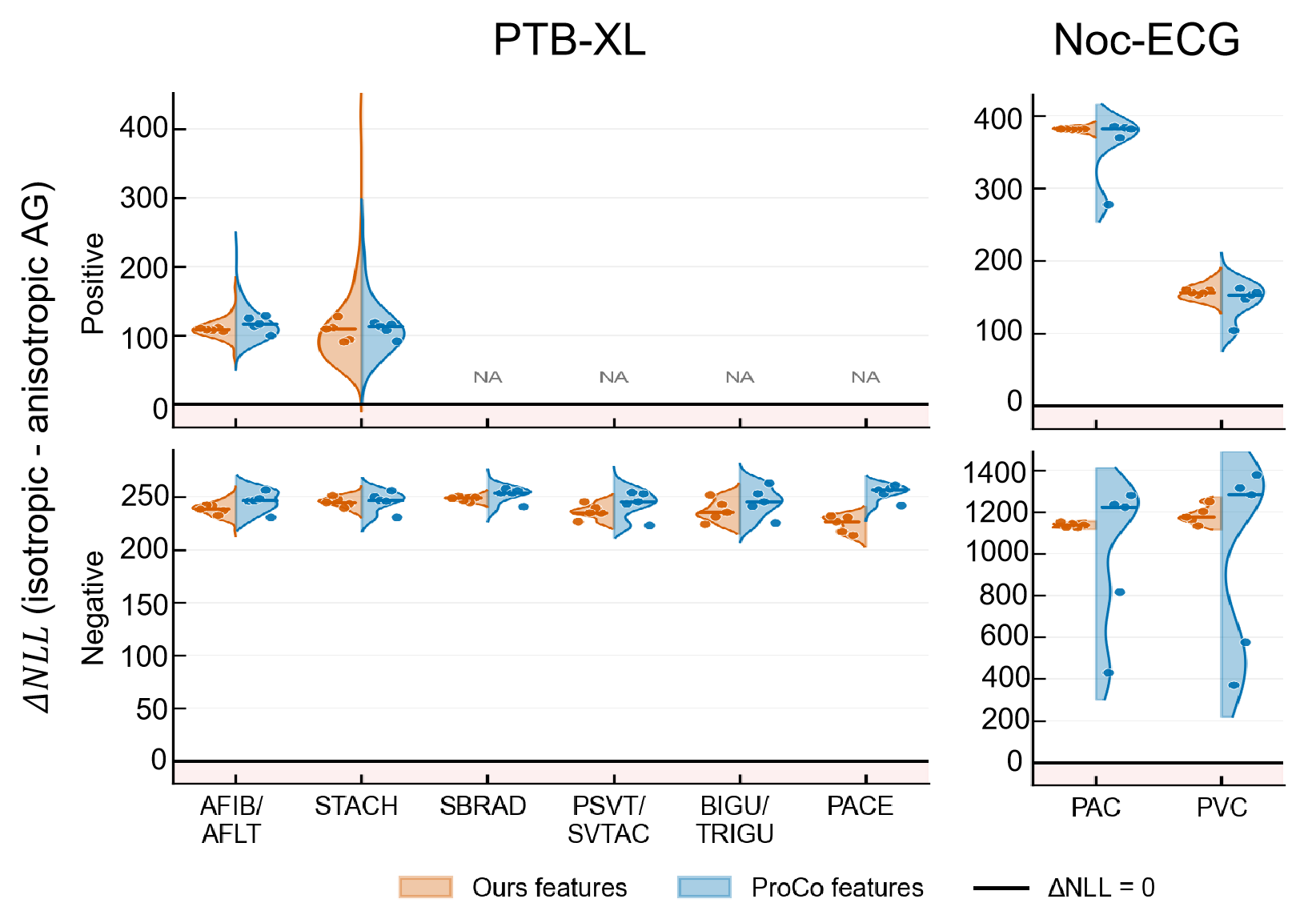}
\caption{Held-out tangent-space fitting supports anisotropic covariance. 
The y-axis reports $\Delta\mathrm{NLL}=\mathrm{NLL}_{\mathrm{isotropic}}-\mathrm{NLL}_{\mathrm{AG}}$; positive values indicate lower held-out NLL for the anisotropic AG surrogate. 
Orange and blue denote AG-SCL and ProCo features, respectively. Half-violins show split-level values across repeated train/held-out splits, with points and horizontal bars marking seed-level means and their median. 
NA indicates PTB-XL positive states with insufficient samples for split-level covariance fitting.}
\label{fig:tangent_fit}
\end{figure}

As shown in Fig.~\ref{fig:tangent_fit}, $\Delta\mathrm{NLL}$ was positive for all eligible evaluations on both datasets. 
This pattern was observed not only for AG-SCL features, but also for ProCo features, indicating that anisotropy was a shared property of the learned hyperspherical ECG representations rather than a feature geometry created only by AG-SCL. 
The improvement was consistent in PTB-XL negative subsets and in the well-sampled PTB-XL positive subsets, and was particularly large in the high-sample Noc-ECG negative states.


\section{Discussion}

This study suggests that long-tailed ECG diagnosis should be treated as both a prior-frequency problem and a representation problem. 
Across the two datasets, AG-SCL improved low-prevalence rhythm detection under operating points that require high specificity. 
The per-class results further show why macro performance alone is insufficient for this setting: aggregate gains can hide the behaviour of clinically important tail classes.

The results support the use of anisotropic representation modelling for ECG rhythms. 
Re-weighting, re-sampling, and fixed logit adjustment mainly modify the influence of class priors, whereas ECG morphologies can vary along direction-dependent timing, amplitude, and beat-shape axes. 
This makes an isotropic class model restrictive for classes whose feature distributions are compact in some directions and diffuse in others. 
Consistent with this view, removing the full Angular Gaussian covariance reduced performance, especially for sensitivity-related metrics on PTB-XL. 
The held-out tangent-space fitting analysis also favoured anisotropic surrogates across eligible label states and seeds, including features learned by ProCo.

The other AG-SCL components appear to moderate the instability introduced by sparse positive states. 
Count-based shrinkage regularizes covariance estimates when only a few positives are available, and ALA learns bounded label-state-specific prior corrections. 
Band-constrained augmentation further increases tail-label view diversity while protecting the QRS-dominant frequency range. 
This interpretation is consistent with the sensitivity analysis, where the largest hyperparameter instability arose from the sub-1\% PTB-XL categories rather than from the head classes or the Noc-ECG labels.

\begin{figure}[!htbp]
\centering
\includegraphics[width=\textwidth]{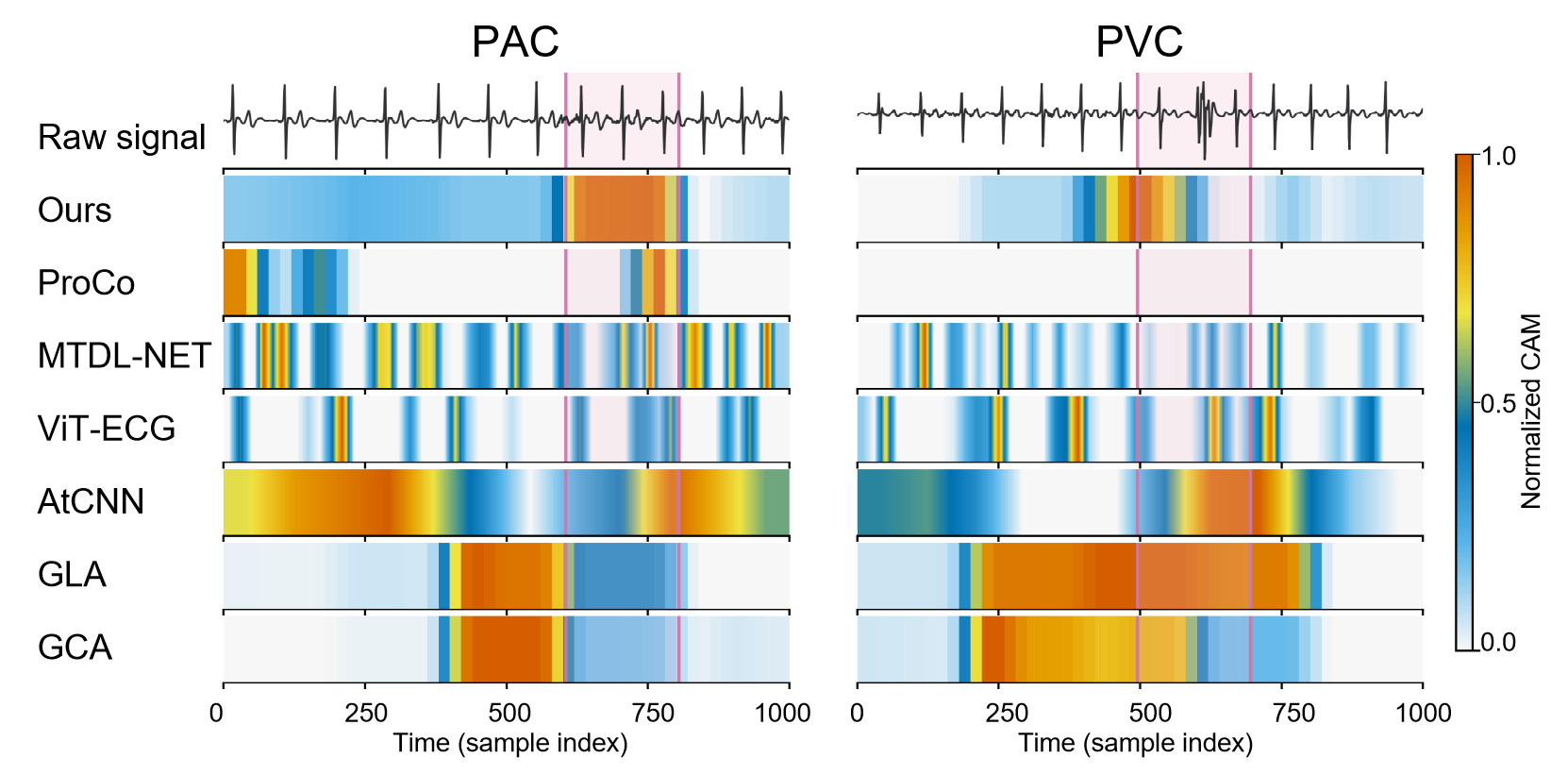}
\caption{Qualitative Grad-CAM illustration on Noc-ECG. Beat-wise annotations are available only for Noc-ECG. Highlighted regions indicate annotated abnormal beats, and heatmaps compare where different methods.}
\label{fig:grad_cam}
\end{figure}

From a clinical-use perspective, the gains in $TPR@FPR_{5\%}$ highlight the relevance of AG-SCL for low-prevalence ECG screening, where improved sensitivity is valuable only when the false-positive burden remains controlled. 
The Noc-ECG evaluation strengthens this point by testing the model on 1{,}317 hours of nocturnal continuous lead-I ECG, a setting closer to long-duration monitoring than short benchmark recordings. 
In this setting, AG-SCL maintained strong detection performance for PAC and PVC, suggesting that its benefits extend beyond curated short-segment datasets. 
Besides, the sensitivity gains over GCA extended across most test subjects, making it less likely that the pooled improvement was driven only by a few high-burden recordings. 
The range-based results also extended the evaluation to temporally contiguous positive segments and showed a clear advantage.
The Grad-CAM examples further provide qualitative context by showing that the model responses often align with annotated abnormal beats in Noc-ECG (Fig.~\ref{fig:grad_cam}).

Several limitations define the scope of these findings. 
Noc-ECG was collected from a single center and from a cohort enriched for neurological disorders or cardiovascular risk, so the results may not fully represent broader screening populations or different acquisition conditions. 
Both datasets were evaluated using lead I, which provides a controlled comparison but does not establish performance for multi-lead inputs or other wearable lead configurations. 
Future work should examine whether these findings generalize across centers, devices, lead configurations, broader screening populations, larger label spaces, and more heterogeneous monitoring cohorts.

\section{Conclusion}

We introduced AG-SCL, a long-tailed ECG arrhythmia framework that combines Angular Gaussian contrastive learning, Adaptive Logit Adjustment, and tail-aware band-constrained augmentation. 
Across PTB-XL and Noc-ECG, AG-SCL improved long-tailed ECG classification, particularly rare-class sensitivity and low-false-positive detection. 
On the held-out Noc-ECG test set, these gains extended across most event-positive subjects, and AG-SCL achieved the highest Range-AUPRC point estimates for temporally contiguous positive segments. 
The ablation results and held-out fitting analysis support anisotropic representation learning as a useful modelling assumption for ECG classes with direction-dependent morphological variation. 
These findings suggest that direction-aware representation learning may improve imbalanced ECG diagnosis, while broader validation remains needed before clinical decision-support use.

\FloatBarrier

\section*{Acknowledgments including declarations}
\subsection*{Statements of ethical approval}
The Noc-ECG cohort was established as a prospective observational study at Shanghai Yangzhi Rehabilitation Hospital, China. The study protocol was reviewed and approved by the Yangzhi Hospital Ethics Committee (Approval No.~(2025)044). Written informed consent was obtained from all participants before ECG acquisition. All Noc-ECG data used in this study were de-identified before analysis. The PTB-XL experiments used a publicly available benchmark dataset, and no additional participant recruitment, participant contact, or access to identifiable information was involved in this secondary analysis.

\subsection*{Funding}
This work was supported in part by the Funding Shanghai Public Health Research Special Project under Grant 2025GKM30, in part by the Tongji University “Medicine + X” Interdisciplinary Research Project under Grant 2025-0648-YB-03, in part by the National Key Clinical Specialty Discipline Construction Project of China under Grant Z155080000004, and in part by the Shanghai Research Center of Rehabilitation Medicine (Top Priority Research Center of Shanghai) under Grant 2023ZZ02027.
\subsection*{Competing interests}
The authors declare that they have no known competing financial
interests or personal relationships that could have appeared to influence
the work reported in this paper.

\section*{Declaration of generative AI and AI-assisted technologies in the writing process}
During the preparation of this work the authors used ChatGPT in order to improve language and readability. After using this tool/service, the authors reviewed and edited the content as needed and take full responsibility for the content of the publication.



\bibliographystyle{elsarticle-num} 
\bibliography{refs}

\end{document}